\documentclass{article}

    \usepackage[final]{tackling_climate_workshop_style}

\usepackage[utf8]{inputenc} 
\usepackage[T1]{fontenc}    
\usepackage{hyperref}       
\usepackage{url}            
\usepackage{booktabs}       
\usepackage{amsfonts}       
\usepackage{nicefrac}       
\usepackage{microtype}      
\usepackage{graphicx}
\usepackage{multirow}
\usepackage{amsmath}
\title{Exploring Variational Graph Autoencoders for Distribution Grid Data Generation}

\author{%
  Syed Zain Abbas \\
  Technical University of Munich \\
  Munich, Germany \\
  \texttt{syedzain.abbas@tum.de} \\
  \And
  Ehimare Okoyomon \\
  Technical University of Munich \\
  Munich, Germany \\
  \texttt{e.okoyomon@tum.de} \\
}

\begin{document}

\maketitle

\begin{abstract}
To address the lack of publicly available power system data for machine learning research in energy networks, we investigate the use of variational graph autoencoders (VGAEs) for generating synthetic distribution grids. Access to detailed feeder models is constrained by privacy and security considerations; yet such data is essential for advancing machine learning approaches that enable reliable and flexible operation of distribution grids with high shares of renewable generation. Using two open-source datasets—ENGAGE and DINGO—we evaluate four decoder variants and compare the generated networks against real grids using structural and spectral metrics. Results show that simple decoders fail to capture realistic topologies, while GCN-based methods achieve strong fidelity on ENGAGE but struggle with the more complex DINGO dataset, producing artifacts such as disconnected components and repeated motifs. These findings highlight both the potential and limitations of VGAEs for grid synthesis, underscoring the need for more expressive generative models and robust evaluation. We release our models and analysis as open source to support benchmarking and accelerate progress in ML-driven power system research.
\end{abstract}

\section{Introduction}
The global transition toward decarbonized energy systems is accelerating as societies work to mitigate climate change. This shift is driving a rapid rise in distributed energy resources (DERs) such as photovoltaics, batteries, and electric vehicles, which are essential for reducing carbon emissions. Their inherently variable and decentralized nature, however, introduces new operational challenges: high renewable penetration can threaten voltage stability and requires distribution networks that are flexible, resilient, and capable of managing rapid fluctuations in supply and demand. Advanced data-driven tools, including machine learning, show promise in supporting reliable, low-carbon grid operation; however, research is constrained by a lack of publicly available distribution grid data. Utilities are reluctant to share feeder topologies and device information due to security and privacy concerns \cite{Mohammadi.2021}. Recent studies emphasize that the quality and coverage of training data directly affect model generalization \cite{Okoyomon_Goebel_2025} and feasibility exploration \cite{Joswig-Jones_Baker_Zamzam_2022}, making this data scarcity a central barrier to developing climate-aligned grid planning and control algorithms.

To address this limitation, synthetic grid generation has emerged as a compelling solution. Rather than relying on sensitive real-world data, researchers construct artificial test systems that emulate key structural and operational properties of distribution networks. Currently, the most widely used approaches for generating this data rely on statistical models and heuristic algorithms \citet{Mohammadi.2021}. Statistical models approximate real-world feeders by sampling from probability distributions of observed grid properties, while heuristic algorithms construct networks by solving optimization problems subject to structural and operational constraints. In practice, both methods induce simplified assumptions that restrict their ability to generate data that captures the diversity, dynamics, and nontrivial relationships that exist in real power systems.

Within the machine learning community, graph-based generative models have gained significant traction in fields such as biology and chemistry \cite{Zhu.13032022}, but relatively few studies have applied these techniques for power systems. Early contributions include FeederGAN \cite{Liang.2021} and DeepGDL \cite{9543363}, whose designs are described in Appendix \ref{app:fgan_dgdl}. However, both face significant issues in their handling of diverse grids and neither model was released to the community for application or benchmarking. This underscores the need for flexible and accessible generative models that are robust and applicable to multiple grid structures and sizes.

This study investigates the use of variational graph autoencoders (VGAEs) as a framework for grid topology generation. Variational autoencoders combine an efficient probabilistic latent representation with flexible decoders that can be tailored to structural constraints, while a graph-based approach allows easy adaptation to various grid topologies. We assess model design across four decoder variants and two distinct datasets. By comparing results on both a small, standardized benchmark and a collection of larger, heterogeneous feeders, we highlight the potential and limitations of using VGAEs to generate realistic topologies. We release our generative models and experimental analysis as open source \footnote{Source code: \url{https://github.com/SyedZainAbbas/GridGEN}} to support the development of new algorithms for distribution network operation and planning, thereby accelerating progress toward a more resilient and sustainable energy system.

\section{Methodology}
We utilize two open-source distribution grid datasets that differ fundamentally in application objective and grid diversity (see Appendix \ref{app:dataset_characteristics}). The ENGAGE dataset \cite{okoyomon_2025_15464235} is a collection of low and medium voltage grids based on the SimBench networks \cite{meinecke2020simbench}, designed to evaluate the generalization capabilities of power flow models across several grid configurations. Due to the small grid sizes and the underlying homogeneity of the SimBench networks, the dataset may not fully capture the diversity found in real-world power systems. The second dataset, DIstribution Network GeneratOr (DINGO), is a large-scale collection of medium voltage feeders with varying sizes, topologies, and structural motifs, built to capture the diversity of realistic distribution systems \cite{article}. DINGO spans a wide size range and includes both simple and complex radial patterns, making it more challenging for generative models to reproduce faithfully.

VGAEs learn a latent representation of network structure through a graph neural encoder and reconstruct graphs via a decoder that predicts edge probabilities \cite{Kipf.21112016}. Four decoder types are considered in this work: (a) a simple inner-product form, (b) an MLP, (c) a Graph Convolutional Network (GCN)-based decoder, and (d) an Iterative-GCN with a refinement loop. Appendix \ref{app:architecture_overview} presents an overview of the architectures of the encoders and decoders used. After training, new synthetic networks are generated by sampling latent variables from the prior distribution and decoding them into adjacency matrices, either in one shot or by using iterative pruning. To evaluate the different decoders, all models are trained on the full datasets.

The models are trained to minimize a variational loss (Appendix \ref{app:loss_function}) that combines reconstruction accuracy with a Kullback–Leibler (KL) divergence term to regularize the latent space \cite{Kipf.21112016}. Evaluation relies on two complementary structural metrics: (average) node degree and the normalized Laplacian spectrum, with similarity quantified using the one-dimensional Wasserstein distance \cite{obray_evaluation_2022}. These jointly capture local connectivity and global structural alignment, providing a balanced measure of model fidelity (see Appendix \ref{app:training_protocol_and_evaluation_metrics}).

\section{Results and Discussion}

\begin{figure}[h]
    \centering
    \includegraphics[width=1.0\linewidth]{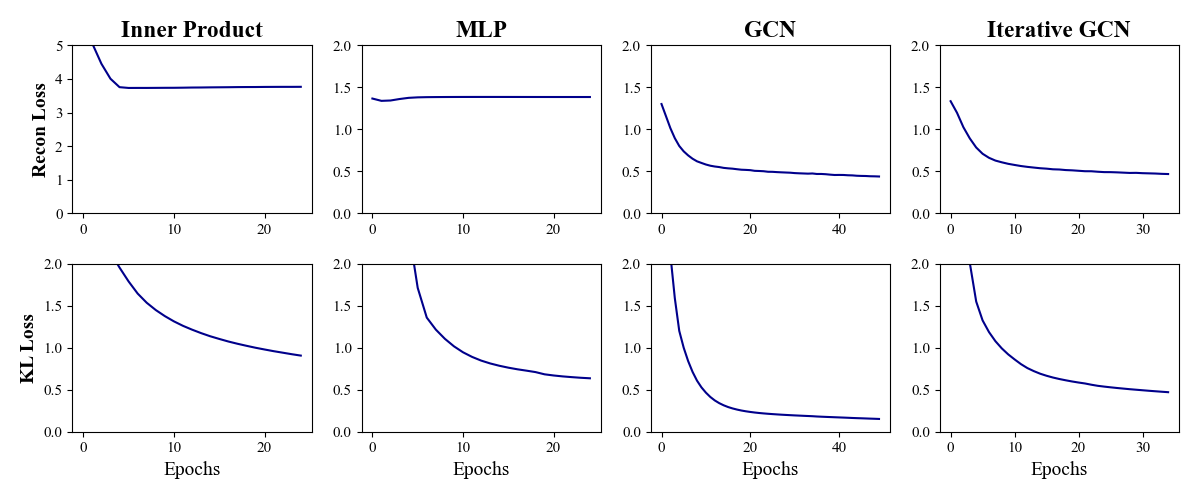}
    \caption{Training loss curves for the four decoder architectures on the DINGO dataset.}
    \label{fig:dingo_losses}
\end{figure}

Figure \ref{fig:dingo_losses} illustrates typical loss curves for each decoder variant. The Inner Product decoder performs poorly on both reconstruction and KL loss, as its simplicity is inadequate for capturing the complexity of grid structures. The MLP decoder shows promising KL loss progression; however, its reconstruction loss plateaus, primarily due to its limited expressive capacity for edge reconstruction in heterogeneous graphs. In contrast, the GCN and Iterative-GCN decoders exhibit similar trends, as their training mechanisms differ only in the graph generation procedure, as outlined in Appendix \ref{app:decoder_architectures}.

Given the weak performance of the Inner Product and MLP decoders during training, they were excluded from subsequent graph generation experiments. Instead, only the GCN and Iterative-GCN decoders were employed to generate synthetic grids. For ENGAGE, 3,000 synthetic graphs were generated, while for DINGO, 1,000 synthetic grids were produced. During generation, the number of nodes was randomly selected within the ranges of 90–100 for ENGAGE and 4,500–7,000 for DINGO, ensuring that the generated graphs were consistent with the scales of the respective training datasets. Across both datasets and under different configurations, the Iterative-GCN decoder consistently outperforms its GCN counterpart, yielding the most accurate structural characteristics.

\begin{table}[htbp]
  \centering
  \caption{Comparison of real and synthetic network properties using the Iterative-GCN model.}
  \begin{tabular}{|l|cc|cc|c|}
    \hline
    \multirow{3}{*}{\textbf{Dataset}} & \multicolumn{4}{c|}{\textbf{Average Degree}} & \textbf{Normalized Laplacian} \\
    \cline{2-5}
    & \multicolumn{2}{c|}{Real} & \multicolumn{2}{c|}{Synthetic} & Wasserstein \\
    & Mean & Std & Mean & Std & Distance \\
    \hline
    \textbf{DINGO} & 1.9986 & 0.0115 & 2.5300 & 1.4651 & 0.5072 \\
    \textbf{ENGAGE} & 2.0521 & 0.0927 & 2.0697 & 0.3066 & 0.1039 \\
    \hline
  \end{tabular}
  \label{tab:results_dingo_vs_engage}
\end{table}

\begin{figure}[h]
    \centering
    \includegraphics[width=1.0\linewidth]{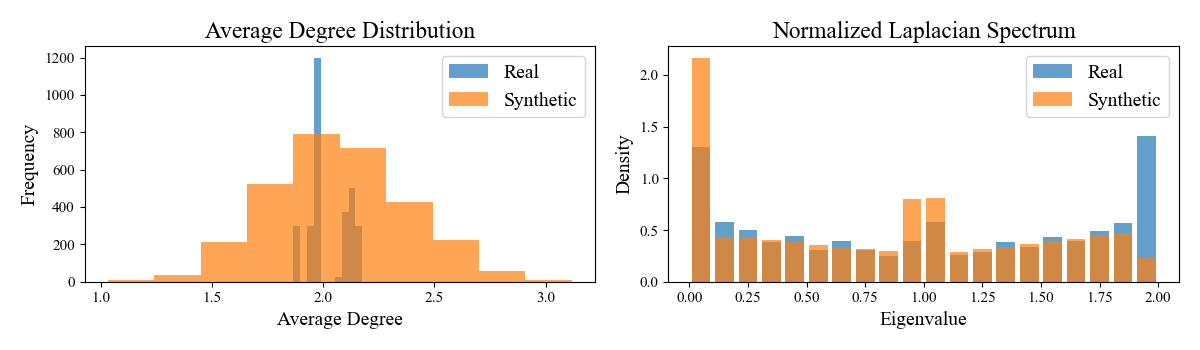}
    \caption{Topological comparison of real and synthetic networks, trained on the ENGAGE dataset.}
    \label{fig:metrics_engage}
\end{figure}

\begin{figure}[h]
    \centering
    \includegraphics[width=1.0\linewidth]{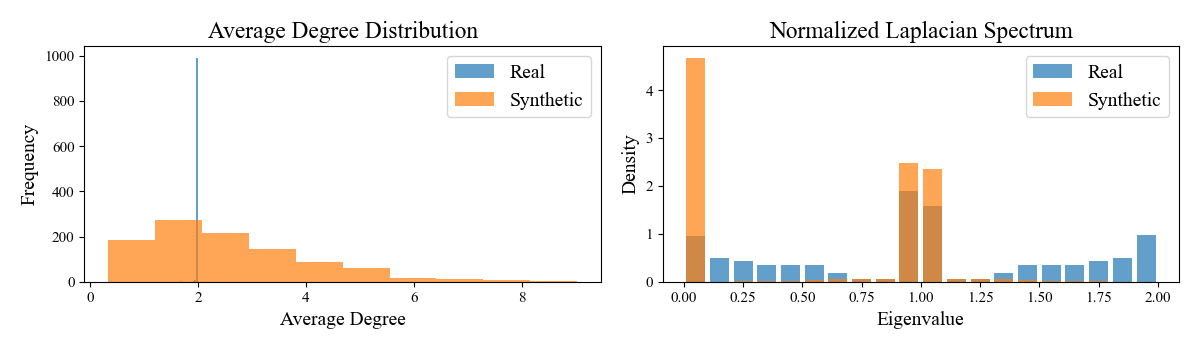}
    \caption{Topological comparison of real and synthetic networks, trained on the DINGO dataset.}
    \label{fig:metrics_dingo}
    
\end{figure}

The results of the real (training) and synthetic grid comparison are summarized in Table \ref{tab:results_dingo_vs_engage}. The ENGAGE-trained synthetic grids closely reproduce the average degree of real networks, indicating that the generative models can match basic radial structure on small, homogeneous benchmarks. On DINGO, however, synthetic graphs display a substantially broader degree distribution, failing to reproduce the average node degree most typical of real feeders. The 1-D Wasserstein distance enables us to perform a spectral comparison of the two datasets based on the normalized Laplacian, with lower distances indicating better global structural alignment. ENGAGE exhibits strong spectral agreement, whereas DINGO shows a large spectral gap, as illustrated in Figures \ref{fig:metrics_engage} and \ref{fig:metrics_dingo}. The DINGO spectra reveal two recurring failure modes in synthetic graphs: (1) an excess of near-zero eigenvalues, consistent with weakly connected or disconnected components, and (2) an over-concentration of mid-range eigenvalues, consistent with repeated artificial motifs.

The performance gap on DINGO versus ENGAGE is driven primarily by dataset complexity. ENGAGE consists of smaller, more homogeneous graphs that present a relatively low-entropy target distribution; in this regime, VGAE priors and decoders can capture both local and global signatures. In contrast, DINGO spans a wide size range and diverse topologies, yielding a distribution that is highly challenging for the VGAE latent prior and the expressive power of the decoders studied. This mismatch implies that success of previous work on specific, smaller grid configurations does not automatically translate to realism for operationally relevant, large-scale feeders. Therefore, evaluation should explicitly test transfer from bench-scale to realistic collections.

These findings have practical implications for the responsible use of machine learning for synthetic grids. Since models trained on datasets that deviate structurally from real networks risk producing brittle or misleading downstream conclusions, spectral and connectivity diagnostics should accompany any downstream benchmark to reveal distributional mismatch. Key limitations of the present work are in the diversity and scale of training data (with DINGO revealing the limits of model coverage), the expressiveness of the VGAE prior and decoders for large graphs, the sensitivity of generation heuristics (since thresholds and retention rules significantly affect outputs), and the narrowness of evaluation metrics (no power-flow feasibility assessment or operational constraints). 

\section{Conclusion}
This study addresses a growing obstacle in machine learning research for power systems: access to realistic distribution grid data. Since utilities are often reluctant to share feeder models due to security and privacy concerns, synthetic network generation emerges as a promising alternative, offering privacy-preserving, representative topologies on which new algorithms can be trained. Within this context, we explore VGAEs as a generative framework for producing realistic distribution grids. We evaluate four decoders across two contrasting datasets and our experiments reveal the reliance of generative model structural fidelity on both architectural design and training data. While basic topological and spectral properties can be recovered on smaller, more homogeneous networks, the same models struggle to reproduce the diversity and complexity of realistic large-scale feeders. Among the architectures, the Iterative-GCN emerged as the most effective decoder, as its refinement mechanism helps enforce sparsity and reduce unrealistic dense structures that appear in other approaches. Future work will combine richer generative families (such as attention-based decoders, diffusion models, or hierarchical approaches) with physics-aware constraints that embed operational feasibility such as loading conditions and power flow convergence into the generation process.
By enhancing data generation methodologies to reflect the diversity of real-world feeders, we can accelerate research on distributed energy resource integration and grid resilience, ultimately enabling data-driven solutions that contribute directly to the decarbonization of power systems.

\appendix
\section{Deep Graph Generation Models for Power Systems}\label{app:fgan_dgdl}
\subsection{FeederGAN \cite{Liang.2021}}
FeederGAN employs a Wasserstein Generative Adversarial Network (WGAN) for power grid generation. Unlike traditional GANs, WGANs leverage the Wasserstein distance as a loss function, which provides more stable gradients during training. In a GAN framework, two neural networks, the generator and the discriminator, engage in a competitive process. The generator aims to produce synthetic data that is indistinguishable from real data, while the discriminator seeks to differentiate between real and synthetic data.
In FeederGAN, this generator is a fully connected multi-layer perceptron (MLP). To train its generator, FeederGAN adopts a device-as-node representation, where devices (e.g., transformers, lines) are modeled as nodes, but bus features are not included. While this allows for simpler graph structures, it may result in the loss of critical node-level features (such as bus features) that are essential for power flow analysis.
FeederGAN faces significant issues related to mode collapse, where the generator produces repetitive topologies or phase configurations. It is also limited to generating only radial networks, making it unsuitable for more complex meshed grid structures, and it achieves a low feasibility ratio of only 27\% with respect to power flow constraints. Furthermore, its reliance on a fully connected MLP restricts its ability to generate graphs with variable sizes and limits its adaptability to larger grid topologies.

\subsection{Deep Graph Distribution Learning (DeepGDL) \cite{9543363}}
DeepGDL offers a more structured approach to power grid generation by utilizing a Recurrent Neural Network (RNN), specifically a Gated Recurrent Unit (GRU), for graph generation. DeepGDL first divides the power grid into subgraphs or communities using a modularity optimization algorithm. Each community is then treated as a smaller subgraph, which is generated separately before being reassembled into a complete grid. On a higher level, the algorithm works like this;

\begin{itemize}
    \item \textit{Community Detection}: The original grid is broken down into smaller, densely connected subgraphs called communities.
    \item \textit{Node-by-Node Generation}: For each community, node and edge features are generated iteratively using the GRU. The process starts with a prior, and at each step, the GRU produces a graph state $S_{i}$, which encodes information about the subgraph generated so far. This state is used as input to two neural networks that generate node features and edge attributes.
    \item \textit{Grid Reconstruction}: Once all nodes in a community are generated, the communities are merged using graph-theoretic techniques to form a single power grid. This merging process often relies on heuristics, which may introduce biases or suboptimal connections.
\end{itemize}

DeepGDL approach of generating subgraphs (communities) individually and combining them into a complete grid provides modularity. However, this process is heavily dependent on the accuracy of community detection, which is heuristic-based. Errors during community detection propagate throughout the entire generation process, potentially affecting the overall grid structure. The use of a RNN-based approach introduces scalability issues, as RNNs are known to struggle with larger graphs due to vanishing gradient problems and increased computational complexity. Moreover, DeepGDL validates the generated grids using graph-theoretic metrics like clustering coefficients and degree distributions, but it does not explicitly guarantee power flow feasibility.

\section{Dataset Characteristics Analysis}\label{app:dataset_characteristics}
\begin{figure}[h]
    \centering
    \includegraphics[width=1.0\linewidth]{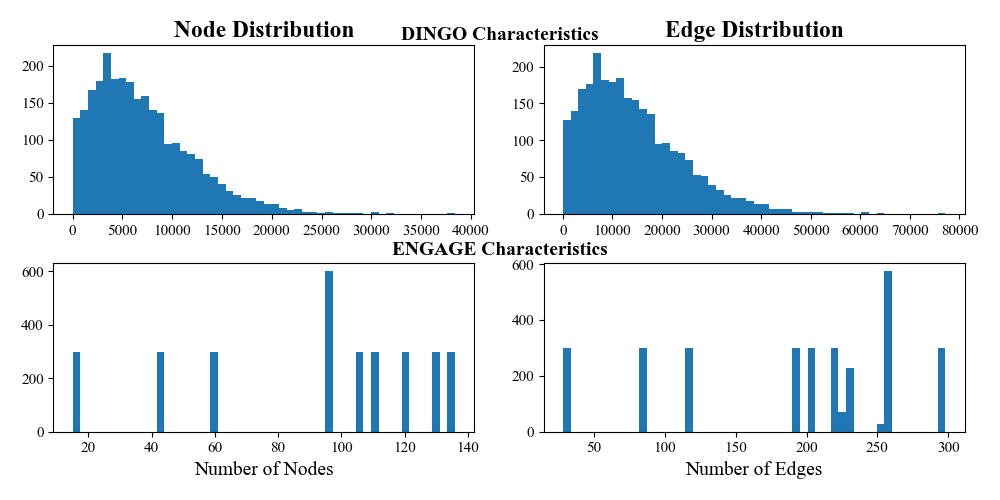}
    \caption{Dataset characteristics comparison between DINGO and ENGAGE datasets. The DINGO dataset exhibits a wide distribution of network sizes with up to 40,000 nodes per grid, while ENGAGE shows discrete clustering around specific node counts, reflecting its origins as a manually created benchmark dataset.}
    \label{fig:dataset_characteristics}
\end{figure}

Fig. \ref{fig:dataset_characteristics} illustrates the graph characteristics of the two datasets used to train the models. The models were trained on these datasets separately to evaluate how different data distributions affect model performance and generalization capabilities.

A notable distinction between the datasets is the significant difference in the number of nodes per graph. This disparity arises from the fundamentally different objectives and methodologies in their creation. DINGO aims to create realistic, high-resolution synthetic medium voltage grids covering all of Germany \cite{article}. The resulting dataset contains 2,722 MVGDs \cite{article}, where each Medium Voltage Grid District (MVGD) typically represents a district supplied by a single HV-MV substation. This focus on comprehensive spatial representation necessitates the generation of numerous districts, each with a detailed network structure and consequently a higher number of nodes to capture local infrastructure characteristics.

The ENGAGE dataset was developed to evaluate model generalization and promote research in robust grid planning and operation \cite{Okoyomon_Goebel_2025}. It has 3000 test cases, created using a statistical sampling approach of the 10 distribution grid cases provided by SimBench \cite{meinecke2020simbench}. The SimBench distribution grids were designed to model German distribution networks at different voltage levels, with radial grids present in the low voltage level and ring structures in the medium voltage networks. The LV grids were derived using a clustering approach and data from official statistics published by the German federal statistical office (DESTATIS) and OpenStreetMap. The MV grids were generated manually rather than algorithmically, with emphasis on creating a limited number of representative MV grid classes (rural, semi-urban, urban, and commercial). This approach results in fewer but well-characterized network topologies with a more manageable number of nodes per graph.

This inherent difference in dataset characteristics — DINGO's large-scale, spatially comprehensive representation versus ENGAGE's focused benchmark approach — presents an interesting opportunity to evaluate deep graph generative models' scalability and generalization capabilities across varying network sizes and complexities.

\section{Architecture Overview}\label{app:architecture_overview}
\subsection{Encoder Architecture}
\begin{figure}[h]
    \centering    
    \includegraphics[width=1.0\linewidth]{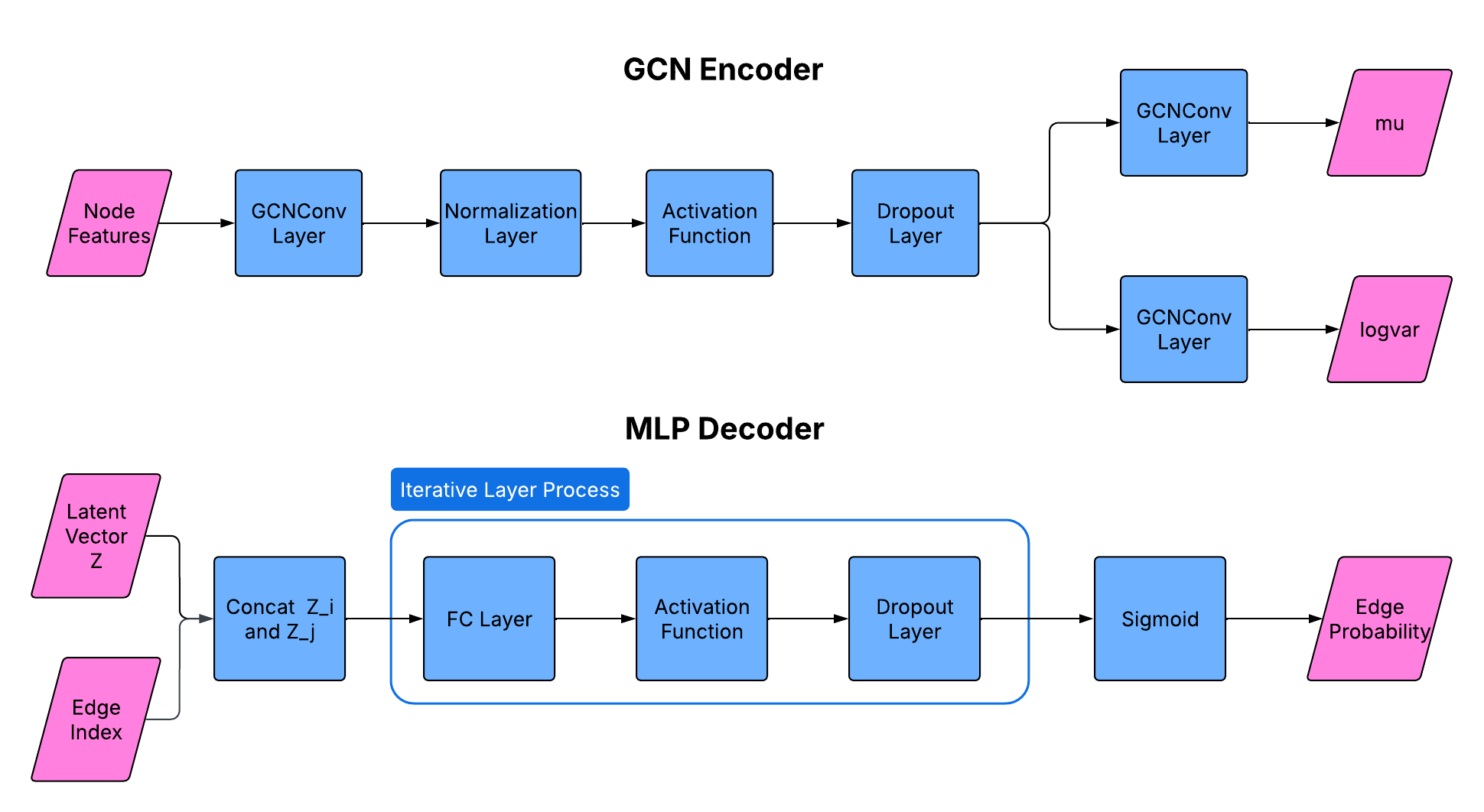}
    \caption{Architecture of the encoder and MLP decoder. The encoder transforms the input graph into a latent representation through GCN layers, normalization, and regularization. The MLP decoder processes concatenated node embeddings through fully connected layers to predict edge probabilities.}
    \label{fig:encoder_mlpdecoder}
\end{figure}

The encoder's primary function is to transform the input graph into a latent representation that can be processed by a decoder, which subsequently learns to reconstruct a topologically similar graph. The architecture employed in this research is illustrated in Fig. \ref{fig:encoder_mlpdecoder}. The input graph is processed through a GCN layer, followed by a layer normalization step to stabilize the learning process. A ReLU activation function is applied to mitigate overfitting, complemented by a dropout layer with a rate of 0.2.

The encoder produces two tensors as output: one representing the mean ($\mu$) of the distribution and another representing the logarithm of the variance ($\log \sigma^2$). The dimensionality of the latent space is a critical hyperparameter that balances representational capacity with computational efficiency. Based on empirical evaluations, a latent dimension of 16 was selected for the ENGAGE dataset. In contrast, a higher dimension of 64 was employed for the more complex DINGO dataset to accommodate its greater topological diversity.

\subsection{Decoder Architectures}\label{app:decoder_architectures}
Four distinct decoder architectures are implemented and compared in this study, each offering different approaches to reconstructing graph structures from latent representations:

\subsubsection{Inner Product Decoder}
The inner product decoder serves as the default implementation in PyTorch Geometric and was initially presented in the seminal work by Kipf and Welling~\cite{Kipf.21112016}. After obtaining the latent variable $Z$, the objective is to determine the similarity between each row in the latent space (where each row corresponds to a vertex) to generate the reconstructed adjacency matrix. 

The inner product computes the cosine similarity between vectors, providing a distance measure that is invariant to vector magnitude. Consequently, by calculating the inner product between $Z$ and its transpose $Z^T$, the similarity between nodes in the latent space can be learned to predict the adjacency matrix. The edge probabilities $A_{ij}$ between nodes $i$ and $j$ are computed as:

\begin{equation}
A_{ij} = \sigma(z_i^T z_j)
\end{equation}

Where $\sigma$ is the sigmoid activation function, ensuring the output is bounded between 0 and 1, representing a probability. While computationally efficient, this decoder assumes that edge formation is determined solely by the dot product of node embeddings, which may not fully capture the complex interdependencies present in power distribution networks.

\subsubsection{MLP Decoder}
The MLP decoder presents a more sophisticated approach than the inner product decoder, though it employs a similar underlying principle. It evaluates the potential existence of an edge between two nodes by concatenating their respective latent representations and processing this combined vector through multiple fully connected layers. 

The architecture of the MLP decoder is illustrated in Fig. \ref{fig:encoder_mlpdecoder} and can be expressed mathematically as:

\begin{equation}
A_{ij} = \sigma(f_{\text{MLP}}([z_i \parallel z_j]))
\end{equation}

Where $[z_i \parallel z_j]$ represents the concatenation of the latent vectors for nodes $i$ and $j$, and $f_{\text{MLP}}$ denotes the MLP function. A sigmoid activation function is applied to the final layer to predict edge probabilities. ReLU activation functions and dropout layers are incorporated between hidden layers as regularization mechanisms. 

The network depth, controlled by the number of fully connected layers, functions as a hyperparameter and depends on the length of the hidden dimension list. The hidden dimension list specifies the dimensionality of each hidden layer.

\subsubsection{GCN Decoder}
\begin{figure}[h]
    \centering    
    \includegraphics[width=1.0\linewidth]{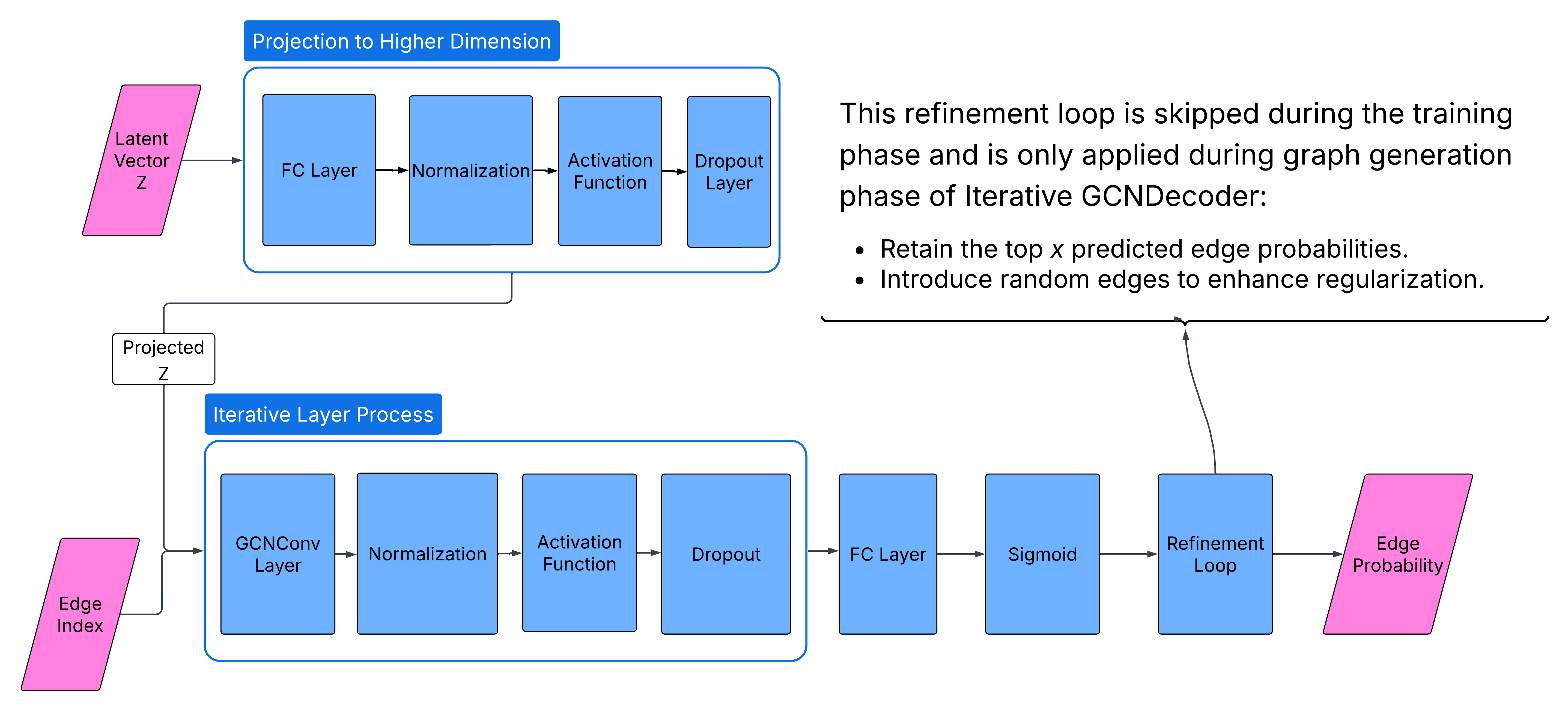}
    \caption{Architecture of the GCN and iterative GCN decoder. The refinement loop, exclusive to the iterative GCN decoder, enhances its graph generation capabilities. The latent representation is processed through projection, normalization, and GCN layers to generate edge probabilities by learning the relationships between nodes in the latent space.}
    \label{fig:gcndecoder}
\end{figure}

This decoder leverages the representational capacity of graph convolutions to predict edge probabilities between node pairs, as illustrated in Fig. \ref{fig:gcndecoder}. Initially, the latent vector is projected into a higher-dimensional space using a linear transformation, followed by normalization, ReLU activation, and dropout operations. Subsequently, this representation is processed through one or more GCN layers. 

The GCN decoder can be mathematically represented as:

\begin{equation}
A = \sigma(f_{\text{GCN}}(Z, \hat{A}))
\end{equation}

Where $Z$ is the latent representation, $\hat{A}$ is an initial edge index (typically randomly generated during the generation phase), and $f_{\text{GCN}}$ represents the graph convolutional function. The number of graph convolution layers is contingent on the complexity of the data.

\subsubsection{Iterative GCN Decoder}
The iterative GCN decoder represents an enhancement of the standard GCN decoder architecture. It introduces three additional hyperparameters specifically for the graph generation phase to address a fundamental challenge in graph generation: standard GCN decoders utilize randomly generated initial edges, which complicates accurate edge probability prediction between nodes and frequently results in excessively dense generated graphs.

The iterative GCN decoder employs the following approach:

\begin{enumerate}
    \item Initial sparse random edge generation using a sparsity constant (typically set between 0.05 and 0.15)
    \item Processing of the latent representation through GCN layers as in the standard GCN decoder
    \item Iterative refinement of edge probabilities through the following process:
\end{enumerate}

For each iteration (the number of iterations being a hyperparameter, typically set to 2):
\begin{itemize}
    \item Retain only the top $x$ edge probabilities (where $x$ is determined by the edge retention ratio hyperparameter, typically between 0.01 and 0.1)
    \item Discard the remaining edges
    \item Add a small number of random edges for regularization (controlled by the exploration edge density hyperparameter, typically between 0.01 and 0.1)
    \item Refine the edge probabilities by passing through a fully connected layer
\end{itemize}

Additionally, an external threshold hyperparameter (not directly part of the model architecture but used during graph generation) determines the cut-off probability for the existence of an edge in the final generated graph. The DINGO dataset typically requires a higher threshold (approximately 0.85) compared to ENGAGE (approximately 0.73), indicating the increased difficulty in creating accurate large-scale graph representations.

The configuration of the best-performing Iterative-GCN decoder for each dataset is presented in Table \ref{tab:training_config}.

\begin{table}[htbp]
  \centering
  \caption{Training configurations for the best performing Iterative-GCN Decoders}
  \begin{tabular}{|l|c|c|c|c|c|}
    \hline
    \textbf{Dataset} & \textbf{Encoder Layers} & \textbf{Decoder Layers} & \textbf{Epochs} &  \textbf{$\beta$ Weight} \\
    \hline
    \textbf{DINGO}  & [128, 64] & [128, 64, 32] & 25 & 2.0\\
    \textbf{ENGAGE} & [64, 16] & [64, 32] & 10 & 5.0\\
    \hline
  \end{tabular}
  \label{tab:training_config}
\end{table}

\subsection{Graph Generation Process}
The practical implementation of graph generation with the VGAE framework requires several modifications from the standard approach. This is particularly necessary because the model expects initial edge indices during the generation phase, and random generation without appropriate constraints produces excessively dense graphs that do not resemble realistic power distribution networks. To address this challenge, several hyperparameters were introduced to control the generation process:

\begin{itemize}
    \item \textbf{Initial edge density:} This parameter establishes the threshold for the edge indices passed to the GCN within the decoder. A higher value produces denser graphs, while a lower value creates more sparse structures. As distribution networks are inherently sparse, empirical testing determined that values between 0.05 and 0.15 yield optimal results, with 0.125 selected as the standard configuration.
    
    \item \textbf{Edge retention ratio:} Utilized specifically in the refinement loop of the iterative GCN decoder, this hyperparameter determines the proportion of predicted edges to retain in each iteration. It effectively selects the top $x$ values from the predicted edge probabilities, with optimal values typically ranging between 0.01 and 0.1, depending on dataset characteristics.
    
    \item \textbf{Exploration edge density:} Also employed in the refinement loop, this hyperparameter functions as a regularizer by introducing random edges to the predicted edge set. This mechanism helps prevent overfitting to local graph structures and promotes diversity in the generated topologies. Empirical testing revealed optimal values between 0.01 and 0.1.
    
    \item \textbf{Threshold:} Implemented as a post-processing step rather than within the model architecture, this parameter provides an additional layer of control over the final graph structure. Predicted edge probabilities must exceed this threshold to be retained in the final graph. The DINGO dataset typically requires a higher threshold (approximately 0.85) compared to ENGAGE (approximately 0.73), reflecting the increased complexity of accurately representing larger network structures.
\end{itemize}

Through careful tuning of these hyperparameters, the VGAE framework can successfully generate synthetic distribution networks that maintain the crucial topological characteristics of real power grids while providing sufficient diversity for comprehensive benchmarking of distribution network algorithms.

\section{Loss Funtion}\label{app:loss_function}

The loss function of a VAE can be formulated as:
\begin{equation}\label{eq:loss_function_vae}
    \mathcal{L}(\theta, \varphi; \mathbf{x}^{(i)}) = -D_{\text{KL}}(q_{\varphi}(z|\mathbf{x}^{(i)}) \parallel p_{\theta}(z)) + \mathbb{E}_{q_{\varphi}(z|\mathbf{x}^{(i)})} \left[ \log p_{\theta}(\mathbf{x}^{(i)}|z) \right].
\end{equation}

The first term is the KL divergence, which quantifies the dissimilarity between two probability distributions. In the context of VAEs, it measures how closely the approximate posterior distribution matches the prior distribution, typically chosen to be Gaussian. This term acts as a regularizer, preventing the encoder from overfitting to the observed data $\mathbf{x}^{(i)}$. The second term represents the reconstruction loss, which evaluates how accurately the model can reconstruct the input data from its latent representation. This paper, \cite{Kipf.21112016}, discusses how to go from a VAE to VGAE.

\begin{figure}[h]
    \centering
    \includegraphics[width=1.0\linewidth]{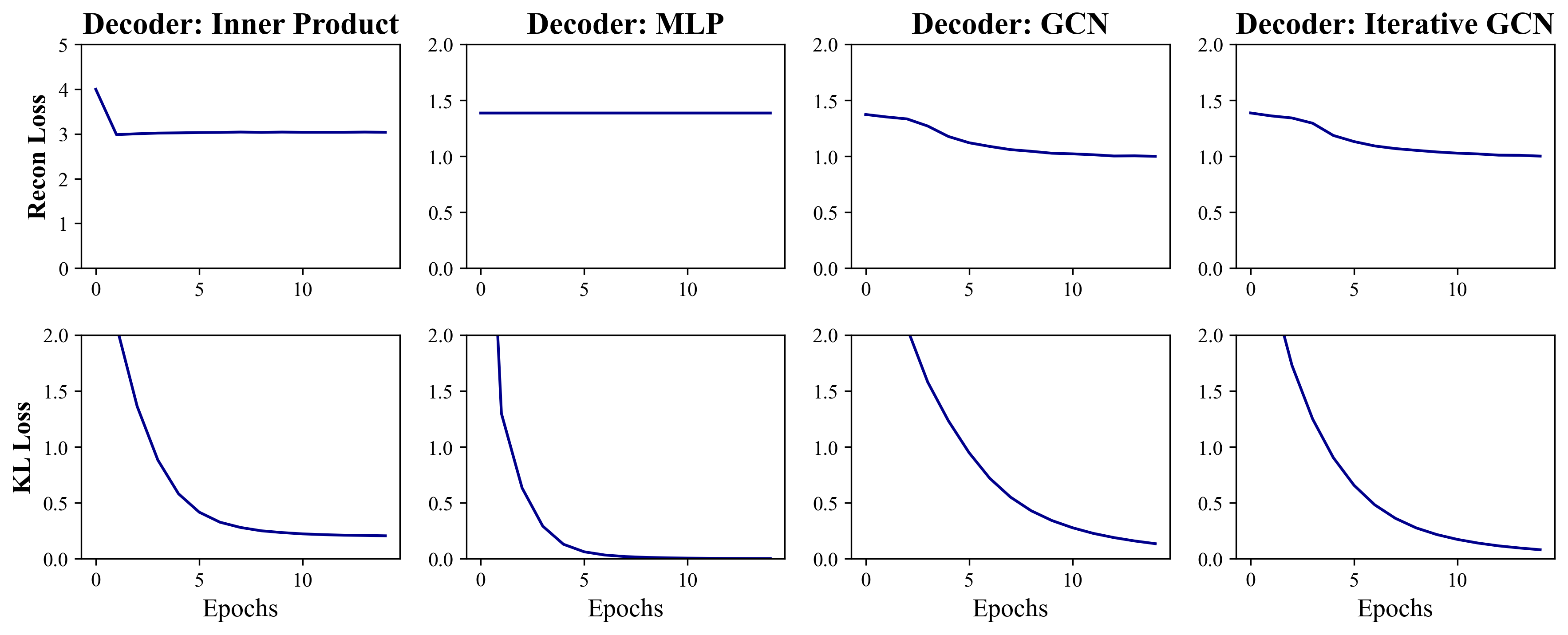}
    \caption{Training loss curves for the four decoder architectures on the ENGAGE dataset.}
    \label{fig:engage_losses}
\end{figure}

The loss curves of the Iterative-GCN are trained on the DINGO and ENGAGE datasets and are presented in Figures \ref{fig:dingo_losses} and \ref{fig:engage_losses}, respectively.


\subsection{Loss Function of VGAE}
The general loss formulation of a VAE (Eq. \eqref{eq:loss_function_vae}) provides the foundation for extending the framework to graph-structured data. In the VGAE setting, this principle is adapted by reconstructing the adjacency matrix of the graph while simultaneously regularizing the latent space. The resulting training objective (Eq. \eqref{eq:loss_function_vgae}) combines a reconstruction loss with a KL divergence term, where a scaling factor $\beta$ allows tuning the trade-off between reconstruction fidelity and latent space regularity.

\begin{equation}\label{eq:loss_function_vgae}
\mathcal{L} = \mathcal{L}_{\text{reconstruction}} + \beta \cdot \mathcal{L}_{\text{KL}}
\end{equation}

Through empirical analysis, it was determined that more complex decoder architectures (GCN and Iterative-GCN) require stronger regularization, with an optimal $\beta$ value of 2 for DINGO and 5 for ENGAGE, while simpler decoders (Inner Product and MLP) performed better with $\beta = 1$.

\section{Training Protocol and Evaluation Metrics} \label{app:training_protocol_and_evaluation_metrics}
All models were trained using the Adam optimizer with a learning rate of 0.001. Due to its smaller size and lower complexity, the ENGAGE dataset required fewer training epochs (typically 10–25), whereas the DINGO dataset needed 25–50 epochs to achieve satisfactory results. A batch size of 32 was used for all models, except for the DINGO dataset when employing complex architectures such as GCN and Iterative-GCN, as these required higher computational power and increased memory usage.

\subsection{Evaluation Metrics} \label{app:evaluation_metrics}
To quantitatively assess the quality of the generated synthetic networks, two primary graph-theoretic metrics were employed:

\subsubsection{Average Degree}
The average degree of a graph provides a fundamental measure of network connectivity and represents the average number of connections per node in the network. For an undirected graph $G = (V, E)$ with $|V|$ nodes and $|E|$ edges, the average degree $\bar{d}$ is calculated as:

\begin{equation}
\bar{d} = \frac{2|E|}{|V|}
\end{equation}

In the context of power distribution networks, average degree carries particular significance as it reflects the redundancy and robustness of the electrical infrastructure. Distribution networks typically maintain relatively low average degrees (between 2 and 3 \cite{PAGANI20132688}) due to their predominantly radial or weakly-meshed topologies, which are designed to balance reliability with economic constraints. Significant deviations in average degree between real and synthetic networks would indicate fundamental structural dissimilarities that could affect the operational characteristics of the modeled system, including fault propagation, power flow patterns, and system restoration capabilities.

The average degree is computationally efficient and intuitively interpretable. However, it provides only limited insight into the global structural properties of the network, necessitating complementary metrics that capture more complex topological structure.
    
\subsubsection{Normalized Laplacian Spectrum}
The normalized Laplacian spectrum provides a sophisticated characterization of a graph's global structure by encoding information about connectivity patterns, clustering tendencies, and community structures. For a graph $G$, the normalized Laplacian matrix $\mathcal{L}$ is defined as:

\begin{equation}
\mathcal{L}_{ij} = 
\begin{cases}
1 & \text{if } i = j \text{ and } d_i \neq 0 \\
-\frac{1}{\sqrt{d_i d_j}} & \text{if } i \text{ and } j \text{ are adjacent} \\
0 & \text{otherwise}
\end{cases}
\end{equation}

where $d_i$ is the degree of node $i$. The eigenvalues of $\mathcal{L}$ form the normalized Laplacian spectrum, which ranges between 0 and 2, with the multiplicity of eigenvalue 0 indicating the number of connected components in the graph. \cite{Chung.1997}

The similarity between the spectra of two graphs can be quantified using some distance metric, like the Wasserstein distance between their eigenvalue distributions. This approach objectively assesses how well a generative model captures the structural properties of reference power grids beyond simple connectivity metrics.

Together, average degree and normalized Laplacian spectrum provide complementary perspectives; the former captures local connectivity patterns while the latter reflects global structural properties. This combination of metrics enables a comprehensive evaluation of how faithfully the proposed deep graph generation methods reproduce the essential topological characteristics of real distribution networks.

These metrics were selected based on their demonstrated ability to capture essential structural characteristics of power distribution networks while remaining computationally tractable for the large number of networks analyzed in this study. We elect not to use the clustering coefficient, a commonly used graph descriptor, as this tends to converge to zero for radial distribution feeders due to limited clustering groups \cite{schweitzer2019creating}.


\begin{thebibliography}{14}
\providecommand{\natexlab}[1]{#1}
\providecommand{\url}[1]{\texttt{#1}}
\expandafter\ifx\csname urlstyle\endcsname\relax
  \providecommand{\doi}[1]{doi: #1}\else
  \providecommand{\doi}{doi: \begingroup \urlstyle{rm}\Url}\fi

\bibitem[Amme et~al.(2018)Amme, Pleßmann, Bühler, Hülk, Koetter, and
  Schwaegerl]{article}
J.~Amme, G.~Pleßmann, J.~Bühler, L.~Hülk, E.~Koetter, and P.~Schwaegerl.
\newblock The ego grid model: An open-source and open-data based synthetic
  medium-voltage grid model for distribution power supply systems.
\newblock \emph{Journal of Physics: Conference Series}, 977:\penalty0 012007,
  02 2018.
\newblock \doi{10.1088/1742-6596/977/1/012007}.

\bibitem[Chung(1997)]{Chung.1997}
F.~R.~K. Chung.
\newblock \emph{Spectral graph theory}, volume no. 92 of \emph{Regional
  conference series in mathematics}.
\newblock {Published for the Conference Board of the mathematical sciences by
  the American Mathematical Society}, Providence R.I., 1997.
\newblock ISBN 0821803158.

\bibitem[Joswig-Jones et~al.(2022)Joswig-Jones, Baker, and
  Zamzam]{Joswig-Jones_Baker_Zamzam_2022}
T.~Joswig-Jones, K.~Baker, and A.~S. Zamzam.
\newblock Opf-learn: An open-source framework for creating representative ac
  optimal power flow datasets.
\newblock In \emph{2022 IEEE Power \& Energy Society Innovative Smart Grid
  Technologies Conference (ISGT)}, page 1–5, Apr. 2022.
\newblock \doi{10.1109/ISGT50606.2022.9817509}.
\newblock URL \url{https://ieeexplore.ieee.org/abstract/document/9817509}.

\bibitem[Khodayar and Wang(2021)]{9543363}
M.~Khodayar and J.~Wang.
\newblock Deep generative graph learning for power grid synthesis.
\newblock In \emph{2021 International Conference on Smart Energy Systems and
  Technologies (SEST)}, pages 1--6, 2021.
\newblock \doi{10.1109/SEST50973.2021.9543363}.

\bibitem[Kipf and Welling(2016)]{Kipf.21112016}
T.~N. Kipf and M.~Welling.
\newblock Variational graph auto-encoders, 2016.
\newblock URL \url{http://arxiv.org/pdf/1611.07308}.

\bibitem[Liang et~al.(2021)Liang, Meng, Wang, Lubkeman, and Lu]{Liang.2021}
M.~Liang, Y.~Meng, J.~Wang, D.~L. Lubkeman, and N.~Lu.
\newblock Feedergan: Synthetic feeder generation via deep graph adversarial
  nets.
\newblock \emph{IEEE Transactions on Smart Grid}, 12\penalty0 (2):\penalty0
  1163--1173, 2021.
\newblock ISSN 1949-3053.
\newblock \doi{10.1109/TSG.2020.3025259}.

\bibitem[Meinecke et~al.(2020)Meinecke, Sarajli{\'c}, Drauz, Klettke, Lauven,
  Rehtanz, Moser, and Braun]{meinecke2020simbench}
S.~Meinecke, D.~Sarajli{\'c}, S.~R. Drauz, A.~Klettke, L.-P. Lauven,
  C.~Rehtanz, A.~Moser, and M.~Braun.
\newblock Simbench—a benchmark dataset of electric power systems to compare
  innovative solutions based on power flow analysis.
\newblock \emph{Energies}, 13\penalty0 (12):\penalty0 3290, jun 2020.
\newblock \doi{https://doi.org/10.3390/en13123290}.

\bibitem[Mohammadi and Saleh(2021)]{Mohammadi.2021}
M.~H. Mohammadi and K.~Saleh.
\newblock Synthetic benchmarks for power systems.
\newblock \emph{IEEE Access}, 9:\penalty0 162706--162730, 2021.
\newblock \doi{10.1109/ACCESS.2021.3124477}.

\bibitem[O'Bray et~al.(2022)O'Bray, Horn, Rieck, and
  Borgwardt]{obray_evaluation_2022}
L.~O'Bray, M.~Horn, B.~Rieck, and K.~Borgwardt.
\newblock Evaluation metrics for graph generative models: Problems, pitfalls,
  and practical solutions.
\newblock In \emph{International Conference on Learning Representations}, 2022.
\newblock URL \url{https://openreview.net/forum?id=tBtoZYKd9n}.

\bibitem[Okoyomon(2025)]{okoyomon_2025_15464235}
E.~Okoyomon.
\newblock Engage dataset: Evaluating network generalization for ac grid
  estimation, May 2025.
\newblock URL \url{https://doi.org/10.5281/zenodo.15464235}.

\bibitem[Okoyomon and Goebel(2025)]{Okoyomon_Goebel_2025}
E.~Okoyomon and C.~Goebel.
\newblock A framework for assessing the generalizability of gnn-based ac power
  flow models.
\newblock In \emph{Proceedings of the 16th ACM International Conference on
  Future and Sustainable Energy Systems}, E-Energy ’25, page 476–488, New
  York, NY, USA, June 2025. Association for Computing Machinery.
\newblock ISBN 979-8-4007-1125-1.
\newblock \doi{10.1145/3679240.3734610}.
\newblock URL \url{https://dl.acm.org/doi/10.1145/3679240.3734610}.

\bibitem[Pagani and Aiello(2013)]{PAGANI20132688}
G.~A. Pagani and M.~Aiello.
\newblock The power grid as a complex network: A survey.
\newblock \emph{Physica A: Statistical Mechanics and its Applications},
  392\penalty0 (11):\penalty0 2688--2700, 2013.
\newblock ISSN 0378-4371.
\newblock \doi{https://doi.org/10.1016/j.physa.2013.01.023}.
\newblock URL
  \url{https://www.sciencedirect.com/science/article/pii/S0378437113000575}.

\bibitem[Schweitzer(2019)]{schweitzer2019creating}
E.~Schweitzer.
\newblock \emph{Creating, Validating, and Using Synthetic Power Flow Cases: A
  Statistical Approach to Power System Analysis}.
\newblock PhD thesis, Arizona State University, 2019.

\bibitem[Zhu et~al.()Zhu, {Du Yuanqi}, Wang, Xu, Zhang, Liu, and
  Wu]{Zhu.13032022}
Y.~Zhu, {Du Yuanqi}, Y.~Wang, Y.~Xu, J.~Zhang, Q.~Liu, and S.~Wu.
\newblock A survey on deep graph generation: Methods and applications.
\newblock URL \url{http://arxiv.org/pdf/2203.06714}.

\end{thebibliography}
\end{document}